\documentclass[11pt,a4paper]{article}
\usepackage[hyperref]{acl2020}
\usepackage{times}
\usepackage{latexsym}
\usepackage{graphicx}
\usepackage{array}
\usepackage{multirow}
\usepackage{rotating}
\usepackage{makecell}
\usepackage{amsmath}
\usepackage{float}

\usepackage{microtype}

\title{Example-Based Named Entity Recognition}

\author{Morteza Ziyadi$^{*}$\\\And Yuting Sun\thanks{* The first two authors contributed equally to this paper}\\\And Abhishek Goswami\\Microsoft Dynamics 365 AI \\
  {\texttt \{moziyadi,yusun,agoswami,jade.huang,wzchen\}@microsoft.com}\\\And Jade Huang\\\And Weizhu Chen \\}
  
\date{}

\begin{document}
\maketitle

\begin{abstract}
We present a novel approach to named entity recognition (NER) in the presence of scarce data that we call example-based NER. Our train-free few-shot learning approach takes inspiration from question-answering to identify entity spans in a new and unseen domain. In comparison with the current state-of-the-art, the proposed method performs significantly better, especially when using a low number of support examples.   

\end{abstract}

\section{Introduction}
Named Entity Recognition (NER) has been a popular area of research within the Natural Language Processing (NLP) community. Most commonly, the NER problem is formulated as a supervised sequence classification task with the aim of assigning a label to each entity in a text sequence. The entity labels typically come from a set of pre-defined categories, such as person, organization, and location. 


A mature technique for handling NER is using expert knowledge to perform extensive feature engineering combined with shallow models such as Conditional Random Fields (CRF) (\citealp{lafferty2001conditional}). The advantages of such a model is that it is easy to train and update the model especially with large datasets available in English (e.g. \citealp{conll2003}), but unfortunately the resulting model is highly associated with known categories, often explicitly memorizing entity values (\citealp{agarwal2020interpretability}). Adding new categories requires further feature engineering, extensive data labeling, and the need to build a new model from scratch or continue to fine-tune the model. This can be challenging, expensive, and unrealistic for real-world settings. For example, when non-expert users create a chatbot in new domains with new business-related entities excluded from conventional NER datasets, they struggle with incorporating these custom entities into their service. 
Additionally, these non-expert users typically have little knowledge of model training, are unwilling to pay for expensive model training services, or do not want to maintain a model. This is very popular for business customers of chatbots such as Alexa and Google Home as well as small business customers of any cloud-based service. 



One existing approach to the challenge of identifying custom entities in a new domain with little data focuses on few-shot learning, which was first designed for the classification task (e.g., \citealp{meta-deng2020}) and recently applied to NER by \citealp{wiseman-stratos-2019-label}, \citealp{zhang2020mzet}, and \citealp{few-shot2018}. The general goal of few-shot learning is to build a model that can recognize a new category with a small number of labeled examples quickly. 
What this means specifically for NER is the training of a model without being constrained by the seen entity types or labels from a source dataset. The model can identify new entities in a completely unseen target domain using only a few supporting examples in this new domain, without any training in that target domain (i.e., train-free).

Existing approaches to few-shot NER have critical limitations despite their successes. For instance,  \citealp{zhang2020mzet} depend on  hierarchical information between coarse-grained and fine-grained entity types to perform few-shot learning and are embedding this crucial hierarchical information into their model architecture. This is hard to generalize to out-of-domain entity types that may not share hierarchical similarities with entities from the training data.  While \citealp{few-shot2018} use a few number of examples to adapt, they never test their model on out-of-domain entities, as the entities in the train, validation, and test set are in fact from the same dataset and following the same distribution.  \citealp{wiseman-stratos-2019-label} propose a train-free approach to adapt to a new domain. But they 
use all of the examples, which number in the thousands, in the target domain to adapt for their evaluation. This is an unrealistic setting, as in many real-world problems, new domains can have as few as 10 or 20 examples available per entity type. We empirically observe that their approach performs poorly in the scenario of using only a few examples as support. 

In this paper, we propose a novel solution to address previous limitations in the train-free and few-shot setting, in which the trained model is directly applied to identify new entities in a new domain without further fine-tuning. The proposed approach is inspired by recent advances in extractive question-answering models \cite{rajpurkar-etal-2018-know} and few-shot learning in language modeling \cite{raffel2019exploring, brown2020language}. First, we formulate  train-free few-shot NER learning as an entity-agnostic span extraction problem with the capability to distinguish sentences with entities from sentences without entities. Our proposed approach is designed to model the correlation between support examples and a query. This way, it can leverage large open-domain NER datasets to train an entity-agnostic model to further capture this correlation. The trained model can be used to perform recognition on any new and custom entities. Second, our model applies a novel sentence-level attention to choose the most related examples as support examples to identify new entities. Third, we systematically compare various self-attention-based token-level similarity strategies for entity span detection. 

We conduct extensive empirical studies to show the proposed approach can achieve significant and consistent improvements over previous approaches in the train-free few-shot setting. For instance, we train a model on the OntoNotes 5.0 dataset and evaluate it on multiple out-of-domain datasets (ATIS, MIT Movie, and MIT Restaurant Review), showing that the proposed model can achieve $>$30$\%$ gain on F1-score using only 10 examples per entity type in the target domain as support, in comparison to \citealp{wiseman-stratos-2019-label}. In addition, we investigate the domain-agnostic properties of different approaches: how much knowledge can be transferred from one domain to another from a similar or different distribution. For instance, in an experiment testing knowledge transfer from a similar domain on the SNIPS dataset, we achieve 48$\%$ gain on F1-score (from 30$\%$ in \citealp{wiseman-stratos-2019-label} to 78$\%$ using our approach) on the SNIPS GetWeather domain using only 10 support examples per entity. For knowledge transfer to a faraway domain, we train a model on OntoNotes 5.0 and run train-free few-shot
evaluation on different mixed datasets and achieve 
significant gain on F1-score. Finally, we perform an ablation study to compare the performance of different training and scoring algorithms for train-free few-shot NER.  


\section{Train-Free Example-Based NER}

In general, the goal of example-based NER is to perform entity recognition after utilizing a few examples for any entity, even those previously unseen during training, as support. 
For example, given this example of the entity \textit{xbox game}, ``I purchased a game called NBA 2k 19" where \textit{NBA 2k 19} is the entity, the \textit{xbox game} entity \textit{Minecraft} is expected to be recognized in the following query ``I cannot play Minecraft with error code 0x111". This simple example demonstrates a single example per single entity type scenario. In real-world scenarios which we have considered in this paper, there are multiple entity types and we have a few examples per entity type. 

\begin{figure*}[h!]
\centering
\includegraphics[width=\textwidth,height=8cm]{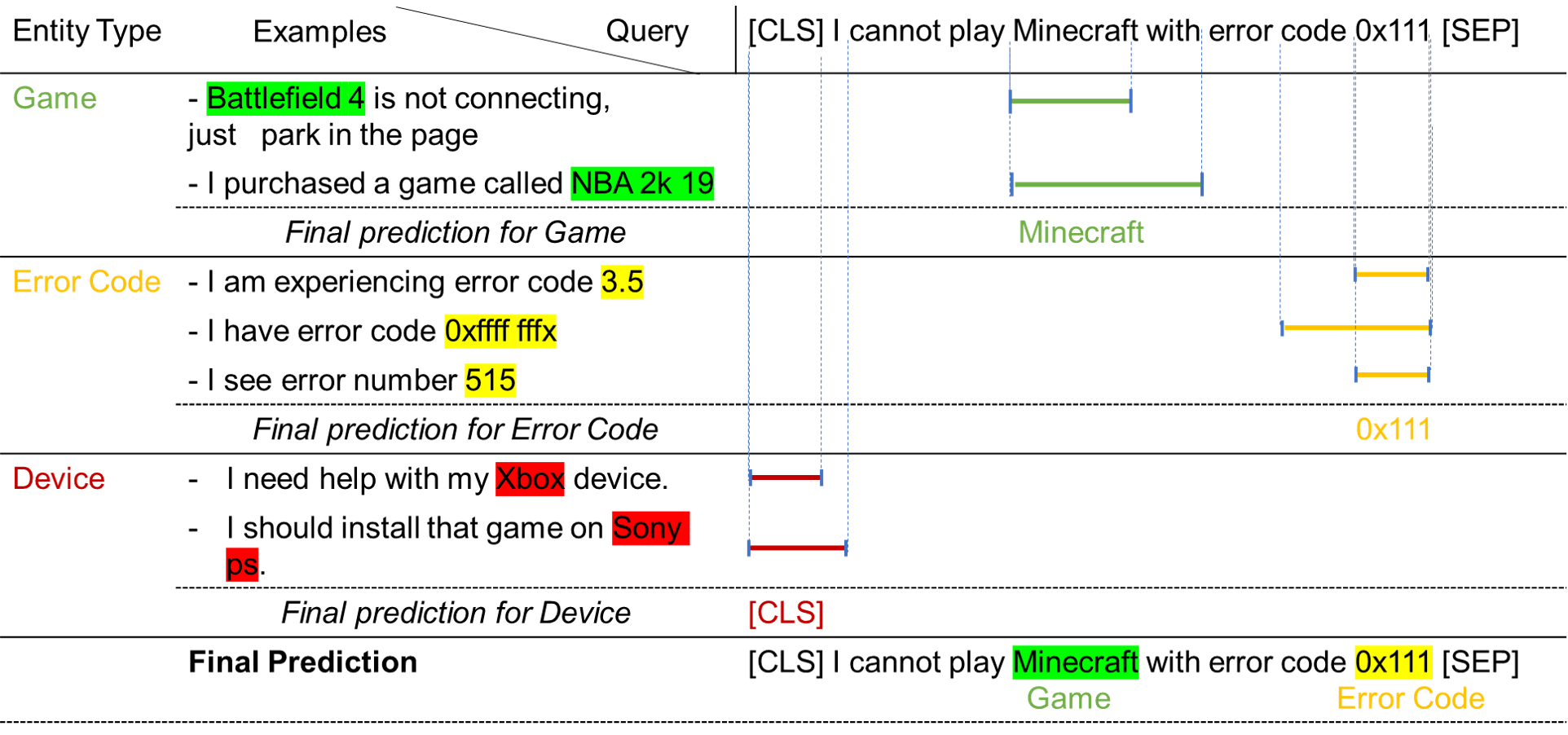}
\caption{An example of example-based NER approach. Prediction per entity type is based on start/end scores of each prediction. Final prediction is an aggregation of the predictions based on the final span scores. For non-entity prediction, the prediction span happens on [CLS] token.}
\label{fig:examples}
\end{figure*}

Figure \ref{fig:examples}
shows an example of example-based entity recognition using a few examples per entity type. In this example, we have two support examples for the ``Game", and ``Device" entity types and three support examples for the ``Error Code" entity type. The goal is to identify the entities in the query ``I cannot play Minecraft with error code 0$\times$111". For each support example for an entity type, we perform span prediction on the query for that entity type and utilize the start/end span scores from each of the predictions to inform the final prediction per entity type. Then, we aggregate the results from different entity types to obtain the final identification of entities in the query. It should be noted that for the aggregation, we also consider the span score which will be explained in detail later. 

There are several challenges that our approach faces. One is how to use multiple examples with different entity types to run train-free scoring. One might consider heuristic voting algorithms as an initial approach but we found that they did not lead to good performance. Another is that we need to fine-tune the language model to get a better representation that can be utilized for a better train-free inference. And finally we have to deal with the gap between the training approach and inference technique. In this paper, we address these challenges and propose a multi-example training and inference approach with a novel attention schema that results in better performance on multiple experiments. 

In order to perform train-free few-shot learning, we fine-tune a BERT language model on a source dataset using a novel similarity-based approach that is independent of seen entity types. We utilize  token-level similarity with sentence-level attention to train the model to produce entity representations. We then use this model to perform prediction in a new domain with several new entity types where each has a few representative examples, such that given a new text, we need to identify whether there is any entity from the set of new entity types in the text. In this setting, the user is able to  define or remove entity types without the need to retrain the model. 

In the following, we first explain our model architecture and then introduce our training and scoring approach. 

\subsection{Model Architecture}
Our approach to solve the problem of recognizing unseen entity types consists of two parts: 1) identifying the span and 2) assigning a specific entity type to each of the detected spans. Span detection aims to predict the start and end of span positions. With the span in hand, we then try to recognize the entity type. Following is a detailed explanation of the architecture. 

\begin{figure*}[h!]
\centering
\includegraphics[width=\textwidth,height=8cm]{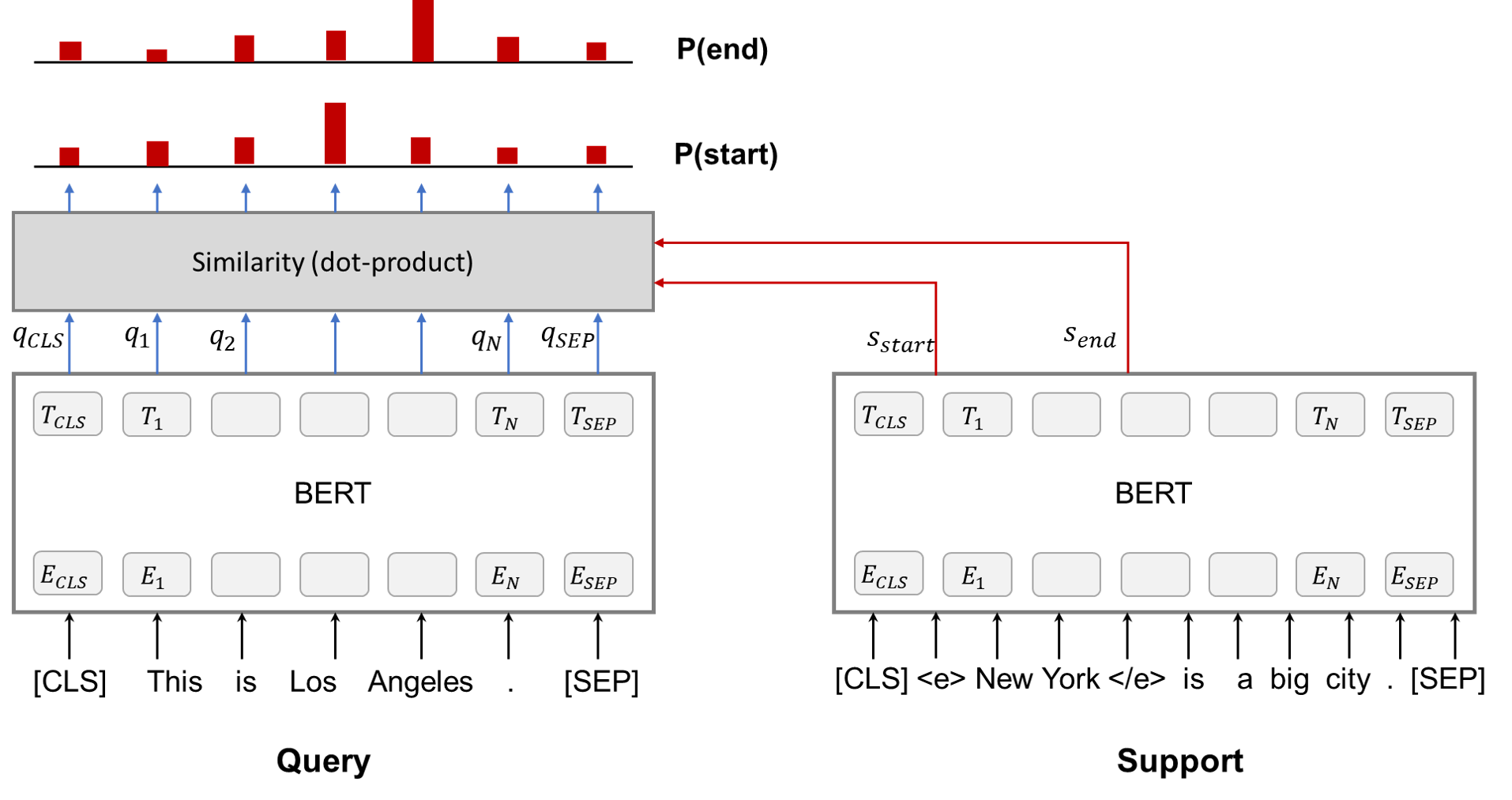}
\caption{The framework of example-based NER}
\label{fig:framework}
\end{figure*}

Figure \ref{fig:framework} shows the framework for the span detection portion of our proposed example-based NER system. Following the overview in Fig. \ref{fig:framework}, we use a similarity-based approach to identify the spans. As shown in the figure, we consider two sets of data: query and support. Query examples are the sentences in which we aim to find the entities, and support examples refer to example sentences with labeled entity types. We  highlight the entities in the support by adding the tokens $<e>$ and $</e>$ to the boundaries of the entities. We leverage the pre-trained language model, BERT, to obtain context-dependent information for each token. 

First, we send the query and support examples separately through the same BERT language model to get an encoded contextual vector for each of the tokens (i.e, $q_i$ for query tokens and $s_i$ for support tokens).

\begin{equation}
\label{eq:1}
 q_i = BERT(w_i^{Query}) 
\end{equation}
\begin{equation}
\label{eq:2}
s_i = BERT(w_i^{Support}) 
\end{equation}

Next, we calculate the start and end probabilities for each of the tokens in the query by measuring the similarity between  $q_i$ and the encoded vectors of $<e>$, $</e>$ tokens in support examples (i.e.,  $s_{start}$, $s_{end}$) which are the boundary vectors of the entities.

\begin{equation}
\label{eq:3}
s_{start} = BERT(w_{<e>}^{Support}) 
\end{equation}
\begin{equation}
\label{eq:4}
s_{end} = BERT(w_{</e>}^{Support})
\end{equation}

To measure similarity, we simply apply the dot-product function between the vectors.
\begin{equation}
\label{eq:5}
{sim}_i^{start} = q_i \odot s_{start} 
\end{equation}
\begin{equation}
\label{eq:6}
{sim}_i^{end} = q_i \odot s_{end}
\end{equation}

The result of the operations so far is the similarity of each of the query tokens with the start/end of an entity type of a single support example. Ideally, we have multiple support examples (e.g., $K$) per entity type. To measure the probability of a token in the query being the start/end of the entity type using multiple support examples, we use the following formula:

\begin{equation}
\label{eq:7}
P_i^{start} = \sum_{j=1}^{K} atten(q_{rep},   s_{rep}^{j})(q_i \odot s_{start}^{j})
\end{equation}
\begin{equation}
\label{eq:8}
P_i^{end} = \sum_{j=1}^{K} atten(q_{rep},   s_{rep}^{j})(q_i \odot s_{end}^{j})
\end{equation}

where $q_i$ is the embedding of token $i$ in the query, $s_{start}^{j}$ $s_{end}^{j}$ are the embeddings of the start and end of an entity respectively in a support example $j$, and $K$ is the number of support examples for that specific entity type. $q_{rep}$, $s_{rep}^{j}$ are the representations of the query and support example $j$, respectively. To calculate the representation of the query and support example, we use the vector sum of all token embeddings in the query and support examples, i.e.,:

\begin{equation}
\label{eq:9}
q_{rep} = VectorSum_{i}(q_{i})
\end{equation}
\begin{equation}
\label{eq:10}
s_{rep}^{j} = VectorSum_{i}(s_{i}^{j})
\end{equation}

The vector sum of all token embeddings of a sentence represents that sentence, either for a query or a support example. 
Another important factor in the above equation is the $atten$ function. We use the following soft attention mechanism to measure sentence-level similarity.

\begin{equation}
\label{eq:11}
atten(q_{rep},   s_{rep}^{j}) = Softmax(T * cos(q_{rep},   s_{rep}^{j}) )
\end{equation}

where $T$ is a hyper-parameter (temperature of $Softmax$ function) and $cos$ is the cosine similarity measure. 
The $atten$ function measures the sentence-level similarity between the query and the support example. We combine this sentence-level similarity with token-level similarity to produce the probability of a token being the start/end of an entity type. We utilize these probabilities to fine-tune  the language model described in the following section.

\subsection{Training}
\label{ssec:layout}

Our approach focuses on fine-tuning the BERT language model in a way such that the contextual encoded text can be used directly for entity recognition. 
We initialize the language model with pre-trained BERT and then pass the minibatches of data to minimize the loss function. Our loss function is the average of span start prediction loss and span end prediction loss using cross-entropy, as follows.  
\begin{multline}\label{eq:12}
 L_{start}=-\sum_{t=1}^{k}(y_t^{start}log{P_t^{start}}\\
 +(1-y_t^{start})log(1-P_t^{start}))
\end{multline}
\begin{equation}
\label{eq:13}
L_{end}=-\sum_{t=1}^{k}(y_t^{end}log{P_t^{end}}+(1-y_t^{end})log(1-P_t^{end}))
\end{equation}
\begin{equation}
\label{eq:14}
 Loss = ( L_{start} + L_{end} ) /2 
\end{equation}

in which k is the length of query; $y_t^{start}$, $y_t^{end}$ indicate whether the token $t$ in the query is the start or end of the entity span, respectively; $P_t^{start}$, $P_t^{end}$ correspond to the probabilities that we calculated during span detection. 

While preparing data, we add $<e>, </e>$ tokens to the support examples around the entities.  We convert a sentence with multiple entities into multiple examples such that each of the support examples include exactly one single entity.
Another important consideration is to construct negative examples as well as positive examples. For instance, for a query ``This is Los Angeles" where ``Los Angeles" is a \textit{city} entity type, the support example ``$<e>$ New York $</e>$ is a big city ." is considered a positive example since it has the same entity type as \textit{city}. But the support example ``Tomorrow, I have an appointment at $<e>$ 2 pm $</e>$." is considered a negative example for the query as it contains a different entity type (in this case \textit{date}). 
In other words, we treat negative examples as no-entity in Eq. \ref{eq:13}. For each training data point, we construct pairs of query and positive/negative support examples to train the model. It should be noted that the ratio of positive versus negative example is an important factor. 
Furthermore, we use multiple examples per entity type for calculating the aforementioned probabilities as Eq. \ref{eq:7} and Eq. \ref{eq:8}. 

\subsection{Entity Type Recognition: Scoring}
\label{sect:pdf}
The second part of our proposed algorithm concerns assigning an entity type to the potential detected spans. One of the approaches used in literature \citep{tanboundary} is to use a softmax layer on top of multi-layer perceptron (MLP) classifier. This type of structure brings limitations to train-free few-shot learning when trying to recognize unseen entity types. Similar to our training schema, for scoring, we instead measure the probability of each of the query tokens being the start and/or end of an entity type using the examples for that entity type. Let us say that we have $M$ entity types with $m_{E_l}$ support examples per entity type and a query $q$. For each entity type $E$ in $E_1, ..., E_M$, we predict with a corresponding score similarly as in training. For each of the tokens in the query, we calculate the scores of being the start $P_{i}^{start}$ and end of a span $P_{i}^{start}$ as:

\begin{equation}
\label{eq:15}
 P_{i}^{start} = \sum_{top K}(q_i \odot s_{start}^{j}) 
\end{equation}
\begin{equation}
\label{eq:16}
P_{i}^{end} = \sum_{top K}(q_i \odot s_{end}^{j})
\end{equation}

It should be noted that these scores are very similar to the probability measure that we had during training but differ when it comes to attention (i.e, Eq. \ref{eq:7}, \ref{eq:8}). During training, we used a soft attention schema with a fixed number of examples per entity type (i.e., $K$). At inference time, we use a hard attention since we have a varying number of support examples per entity type. For the hard attention, first we calculate the token-level similarity measure $(q_i \odot s_{start/end}^{j})$ for each of the entities in the support examples and then pick the top K ones with highest measure and then sum their similarity scores with an equal attention of 1 for these example, to calculate the start/end probability. In other words, we sum the token-level similarity of the top $K$ support examples with highest $(q_i \odot s_{start/end}^{j})$ for probability calculation. Then, for each potential span, we calculate the total span score as the summation of start and end scores, i.e., $score(span) = P^{start} + P^{end}$ where $P^{start}$ and $P^{end}$ are calculated using above equations (Eq. \ref{eq:15}, \ref{eq:16}). After getting all the potential spans, we select the one with the highest score as the final span for that specific entity type, treating span prediction similarly as question-answering (QA) models.

So far, we have finalized the spans for each of the entity types. Algorithm \href{alg:1}{I} shows the top span prediction per entity type. We should highlight that similar to the QA framework, if the predicted span's start and end occur on the CLS token, we treat it as no span for that entity type in the query. 

\begin{table}[h!]
\centering
\begin{tabular}{p{0.45\textwidth}}
\hline \textbf{Algorithm I: Top span prediction per entity type} \\
\hline
$start_{indexes}$, $end_{indexes}$ = range(len(query($q$)))   \\
\textbf{for} $start_{id}$ in $start_{indexes}$ \textbf{do}: \\
$|$\hspace{10pt} \textbf{for} $end_{id}$ in $end_{indexes}$ \textbf{do}:  \\
$|$ \hspace{10pt} $|$ \hspace{10pt} \textbf{if} $start_{id}$ , $end_{id}$ in the query \\
$|$ \hspace{10pt} $|$ \hspace{20pt}$\&$ $start_{id}$ $<$ $end_{id}$:  \\
$|$ \hspace{10pt} $|$ \hspace{10pt} $|$ \hspace{10pt}calculate $P^{start}, P^{end}$ using \\
$|$ \hspace{10pt} $|$ \hspace{10pt}  \hspace{27pt}Eq. \ref{eq:15}, \ref{eq:16}\\
$|$ \hspace{10pt} $|$ \hspace{10pt} $|$\hspace{10pt} add ($start_{id}$, $end_{id}$,\\
$|$ \hspace{10pt} $|$ \hspace{10pt}  \hspace{27pt}$P^{start}$, $P^{end}$) to the output \\
\\
\textbf{sort} the spans based on the ($P^{start}$ + $P^{end}$) to select the top high probable span \\
\hline
\end{tabular}
\label{alg:1}
\end{table}

To merge the spans from different entity type predictions, we use their score to sort,  remove the overlap, and obtain the final predictions. Algorithm \href{alg:2}{II} shows the overall scoring functionality. 

\begin{table}[h!]
\centering
\begin{tabular}{p{0.45\textwidth}}
\hline \textbf{Algorithm II: Entity Type Recognition - Hard Attention Algorithm} \\
\hline
Suppose we have $M$ entity types with $m_l$ support examples per entity type and a query $Q$.\\
\hline
\textbf{for} each entity type $E$ in $E_1, .., E_M$ \textbf{do}:\\
$|$\hspace{10pt} \textbf{get} the span prediction per entity type using\\
\hspace{30pt}Algorithm \href{alg:1}{I}  \\
\\
\textbf{aggregate} all the predictions per entity type and sort based on span score.\\ 
\textbf{remove overlaps:} select the top score span and search for the second top span without any overlap with the first one and continue this for all the predicted spans.\\
\hline
\end{tabular}
\label{alg:2}
\end{table}

Aside from the hard attention used in our scoring method, we also experimented with other scoring methods which are explained in the ablation study.

\section{Experimental Setup}

\subsection{Dataset Preparation}
We test our proposed approach on different benchmark datasets that are publicly available with results in tables \ref{tab:public_dataset_1} and \ref{tab:public_dataset_2}. The datasets are OntoNotes5.0 \footnote{from: https://github.com/swiseman/neighbor-tagging}, Conll2003 \footnote{from: https://github.com/swiseman/neighbor-tagging}, ATIS \footnote{from: https://github.com/yvchen/JointSLU}, MIT Movie and Restaurant Review \footnote{from: https://groups.csail.mit.edu/sls/downloads/}, and SNIPS \footnote{from:https://github.com/snipsco/nlu-benchmark/tree/master/2017-06-custom-intent-engines}.  Additionally, we use a proprietary dataset (Table \ref{tab:proprietary_dataset_1}) as a test set in some of the experiments.

\begin{table*}[ht!]
\centering
\begin{tabular}{c|c|c|c|c|c}
Dataset & OntoNotes 5.0 & ATIS & Movie.Review & Restaurant.Review & Conll2003\\ 
\hline
$\#$Train(Support) & 59.9k & 4.6k & 7.8k & 7.6k & 12.7k\\ 
$\#$Test & 7.8k & 850 & 2k &1.5k & 3.2k \\ 
$\#$Entity Types & 18 & 79 & 12 & 8 & 4\\ 
\hline
\end{tabular}
\caption{Public datasets statistics for OntoNotes 5.0, ATIS, Movie.Review, Restaurant.Review, and Conll2003}
\label{tab:public_dataset_1}
\end{table*}

\begin{table*}[h]
\centering
\begin{tabular}{c|c|c|c|c|c|c|c}
{\multirow{2}{*}{Dataset}}&\multicolumn{7}{c}{SNIPS}\\\cline{2-8}&d1&d2&d3&d4&d5&d6&d7\\\Xhline{1\arrayrulewidth}
$\#$Train (Support) & 2k & 2k & 2k & 2k & 2k & 2k & 1.8k \\
$\#$Test & 100 & 100 & 99 & 100 & 98 & 100 & 100  \\ 
$\#$Entity Types & 5 & 14 & 9 & 9 & 7 & 2 & 7 \\ \Xhline{1\arrayrulewidth}
\end{tabular}
\caption{Public datasets statistics for SNIPS dataset with different domains of: \textbf{d1}: AddToPlaylist, \textbf{d2}: BookRestaurant, \textbf{d3}: GetWeather, \textbf{d4}: PlayMusic, \textbf{d5}: RateBook, \textbf{d6}: SearchCreativeWork, \textbf{d7}: SearchScreeningEvent}
\label{tab:public_dataset_2}
\end{table*} 

We first fine-tune the language model on source datasets and then run train-free few-shot evaluation on a target domain with unseen entity types. During the fine-tuning, we use all the training data from the source datasets. When predicting an unseen entity in target domain, we sample a subset of instance in the training dataset from the target domain as the support set. We run the sampling multiple times and produce the metric with mean and standard deviation in the experimental results.  Note that none of the examples or entity types of the target dataset are seen in the fine-tuning step. 

\begin{table}[h]
\begin{tabular}{p{0.3\textwidth}|p{0.1\textwidth}}
\centering
Dataset & Proprietary\\ 
\hline
$\#$Train(Support) & 8.6k\\ 
$\#$Test & 3k \\ 
$\#$Entity Types & 86\\ 
$\#$Examples/EntityType in Test & 2 $\sim$ 465\\ 
\hline
\end{tabular}
\caption{Proprietary dataset statistics}
\label{tab:proprietary_dataset_1}
\end{table}

\subsection{Training and Evaluation Details}
We use the PyTorch framework (\citealp{NIPS2019_9015}) for all of our experiments. To train our models, we use the AdamW optimizer with a learning rate of $5e-5$, adam epsilon of $1e-8$, and weight decay of $0.0$. The maximum sentence length is set to $384$ tokens and $K$, the number of example per entity type for training and hard attention scoring, is $5$. For training, $T$, the temperature of attention, is $1$. We use the  BERT-base-uncased model to initialize our language model. For evaluation, we report the precision, recall, and F1 scores for all  entities in the test sets based on exact matching of entity spans. 

\section{Experimental Results}

\subsection{Benchmark Study}

\subsubsection{Knowledge Transfer from a General Domain to Specific Domains}
In this section, we run a benchmark study to investigate the performance of our approach. In the first set of experiments, we investigate how we can transfer knowledge from a general domain to specific domains. We train a model on a general source domain and then immediately evaluate the model on the target domains with support examples. 
Figure \ref{tab:onto_2_other_main} shows the results of training a model on the OntoNotes 5.0 dataset as a generic domain and evaluating it on other target domains (i.e., ATIS, MIT Movie Review, and MIT Restaurant Review, and MixedDomains). We use Neighborhood Tagging \cite{wiseman-stratos-2019-label} as the baseline. It clearly demonstrated that our proposed approach is significantly better than the baseline and the improvements are  consistent in all the three datasets. 
\subsubsection{Performance Using Differing Numbers of Support Examples}
Next, we investigate how different approaches perform with a differing number of support examples per entity type. In these experiments, we use different support samples from the datasets for inference, randomly sampling a fixed number of examples per entity type (i.e., 10, 20, 50, 100, 200, 500 examples per entity type) from the overall support set to use for inference. Note that if an entity has a smaller number of support examples than the fixed number, we use all of them as our support examples. For example, if an entity type has 30 examples in total, for the case of 100 examples per entity type, we use all 30 examples for that entity type. If an entity has 1000 examples, we sample a fraction of it (e.g., 100 for the experiment on 100 examples per entity type). Table \ref{tab:onto_2_other_main} shows this study and we observe that our approach performs much better in comparison to the neighbor-tagging \citep{wiseman-stratos-2019-label} work. These results are based on running the experiments 10 times and taking the average of the F1-score of them. Also, we show the standard deviation for such multiple experiments.

An important point to highlight is the higher performance of our approach especially with a low number of support examples. Even with as few as 10 or 20 examples per entity types, in some datasets (e.g., MIT Movie and Restaurant Review), our approach can achieve results as good as using a higher number of examples per entity types such as 500. 

\begin{table*}[h]
\centering
\begin{tabular}{c|c|c|c|c|c|c|c}
\multirow{2}{*}{Target Dataset}&\multirow{2}{*}{Approach}&\multicolumn{6}{c}{\textit{Number of support examples per entity type}}\\\cline{3-8}
&&\textit{10}&\textit{20}&\textit{50}&\textit{100}&\textit{200}&\textit{500}\\\Xhline{1\arrayrulewidth}
\multirow{2}{*}{ATIS}&Neigh.Tag.& 6.7$\pm$0.8 & 8.8$\pm$0.7 & 11.1$\pm$0.7& 14.3$\pm$0.6& 22.1$\pm$0.6& 33.9$\pm$0.6\\
&\textit{Ours} & \textbf{17.4$\pm$1.1} & \textbf{19.8$\pm$1.2} & \textbf{22.2$\pm$1.1} & \textbf{26.8$\pm$2.7} & \textbf{34.5$\pm$2.2} & \textbf{40.1$\pm$1.0}\\
\Xhline{2\arrayrulewidth}
\multirow{2}{*}{MIT Movie}&Neigh.Tag.& 3.1$\pm$2 & 4.5$\pm$1.9 & 4.1$\pm$1.1 & 5.3$\pm$0.9 & 5.4$\pm$0.7 & 8.6$\pm$0.8 \\
&\textit{Ours} & \textbf{40.1$\pm$1.1} & \textbf{39.5$\pm$0.7} & \textbf{40.2$\pm$0.7} & \textbf{40.0$\pm$0.4} & \textbf{40.0$\pm$0.5} & \textbf{39.5$\pm$0.7} \\
\Xhline{2\arrayrulewidth}
\multirow{2}{*}{MIT.Restaurant}&Neigh.Tag.& 4.2$\pm$1.8 & 3.8$\pm$0.8 & 3.7$\pm$0.7 & 4.6$\pm$0.8 & 5.5$\pm$1.1 & 8.1$\pm$0.6\\
&\textit{Ours} & \textbf{27.6$\pm$1.8} & \textbf{29.5$\pm$1.0} & \textbf{31.2$\pm$0.7} & \textbf{33.7$\pm$0.5} & \textbf{34.5$\pm$0.4} & \textbf{34.6} \\
\Xhline{2\arrayrulewidth}
\multirow{2}{*}{Mixed Domain}&Neigh.Tag.& 4.5$\pm$0.8 & 5.5$\pm$0.5 & 6.7$\pm$0.6 & 8.7$\pm$0.4 & 13.1$\pm$0.6 & 20.3$\pm$0.4\\
&\textit{Ours} & \textbf{16.6$\pm$1.1} & \textbf{20.3$\pm$0.8} & \textbf{23.5$\pm$0.6} & \textbf{27.4$\pm$1.2} & \textbf{32.2$\pm$1.2} & \textbf{35.9$\pm$0.6} \\
\Xhline{1\arrayrulewidth}
\end{tabular}
\caption{General domain (OntoNotes 5.0) to specific target domains: ATIS, Moive Review, Restaurant Review,  MixedDomains (ATIS+MIT.Restaurant.Review as mixed distributed dataset), impact of number of support examples per entity type, comparison of our approach with the Neighbor Tagging \citep{wiseman-stratos-2019-label}} in terms of F1-score (mean$\pm$standard deviation is calculated based on 10 random samples)
\label{tab:onto_2_other_main}
\end{table*}


\subsubsection{Performance When the Target Dataset Contains Different Distributions}

We also analyze the domain-agnostic feature of different approaches, with the goal of finding out how different approaches perform for a scenario where the target dataset includes datapoints from different distributions. This can arise due to different data sources, different criteria for labeling, and/or when new entity types from different domains are added to the system. To analyze such a scenario, we combine the ATIS and MIT Restaurant Review datasets and run evaluation on this new mixed data. The MixedDomains results in Table \ref{tab:onto_2_other_main} shows the performance analysis of the model trained on OntoNotes 5.0 and evaluated on the new mixed domain dataset. Based on this figure, we conclude that our approach achieves the best overall performance (F1-score) under different scenarios of support examples.

\subsubsection{Performance Using Another Source Dataset to Fine-Tune the Language Model}

Besides using OntoNotes 5.0 dataset as a training set, we also train a model on Conll2003 to function as a generic domain dataset and evaluate it in a train-free few-shot manner on other domains. Table \ref{tab:conll_2_other_main} shows similar results as when training with OntoNotes 5.0. When we compare tables \ref{tab:onto_2_other_main} and \ref{tab:conll_2_other_main}, we see that knowledge transfer from OntoNotes 5.0 achieves overall higher performance in comparison to Conll2003. We conjecture this is due to the larger size and larger number of entity types in OntoNotes 5.0, compared to Conll2003.

\begin{table*}[h]
\centering
\begin{tabular}{c|c|c|c|c|c|c|c}
\multirow{2}{*}{Target Dataset}&\multirow{2}{*}{Approach}&\multicolumn{6}{c}{\textit{Number of support examples per entity type}}\\\cline{3-8}
&&\textit{10}&\textit{20}&\textit{50}&\textit{100}&\textit{200}&\textit{500}\\\Xhline{1\arrayrulewidth}
\multirow{2}{*}{ATIS}&Neigh.Tag.& 2.4$\pm$0.5 & 3.4$\pm$0.6 & 5.1$\pm$0.4 & 5.7$\pm$0.3 & 6.3$\pm$0.3 & 10.1$\pm$0.4 \\
&\textit{Ours} & \textbf{22.9$\pm$3.8} & \textbf{16.5$\pm$3.3} & \textbf{19.4$\pm$1.4} & \textbf{21.9$\pm$1.2} & \textbf{26.3$\pm$1.1} & \textbf{31.3$\pm$0.5}\\
\Xhline{2\arrayrulewidth}
\multirow{2}{*}{MIT Movie}&Neigh.Tag.& 0.9$\pm$0.3 & 1.4$\pm$0.3 & 1.7$\pm$0.4 & 2.4$\pm$0.2 & 3.0$\pm$0.3 & 4.8$\pm$0.5  \\
&\textit{Ours} & \textbf{29.2$\pm$0.6} & \textbf{29.6$\pm$0.8} & \textbf{30.4$\pm$0.8} & \textbf{30.2$\pm$0.6} & \textbf{30.0$\pm$0.5} & \textbf{29.6$\pm$0.5} \\
\Xhline{2\arrayrulewidth}
\multirow{2}{*}{MIT.Restaurant}&Neigh.Tag.& 4.1$\pm$1.2 & 3.6$\pm$0.8 & 4.0$\pm$1.1 & 4.6$\pm$0.6 & 5.6$\pm$0.8 & 7.3$\pm$0.5 \\
&\textit{Ours} & \textbf{25.2$\pm$1.7} & \textbf{26.1$\pm$1.3} & \textbf{26.8$\pm$2.3} & \textbf{26.2$\pm$0.8} & \textbf{25.7$\pm$1.5} & \textbf{25.1$\pm$1.1} \\
\Xhline{2\arrayrulewidth}
\multirow{2}{*}{Mixed Domain}&Neigh.Tag.& 2.3$\pm$0.5 & 2.9$\pm$0.5 & 4.1$\pm$0.6 & 4.7$\pm$0.3 & 5.4$\pm$0.4 & 7.9$\pm$0.4\\
&\textit{Ours} & \textbf{20.5$\pm$1.6} & \textbf{18.6$\pm$1.9} & \textbf{20.9$\pm$1.2} & \textbf{22.5$\pm$0.5} & \textbf{24.7$\pm$0.9} & \textbf{27.3$\pm$0.5} \\
\Xhline{1\arrayrulewidth}
\end{tabular}
\caption{General domain (Conll2003) to specific target domains, impact of number of support examples per entity type, comparison of our approach with Neighbor Tagging \citep{wiseman-stratos-2019-label}} in terms of F1-score (mean$\pm$standard deviation is calculated based on 10 random samples)
\label{tab:conll_2_other_main}
\end{table*}


\subsubsection{Knowledge Transfer from a Similar Domain}
\begin{figure*}[h!]
\centering
\includegraphics[width=\textwidth,height=0.4\textwidth]{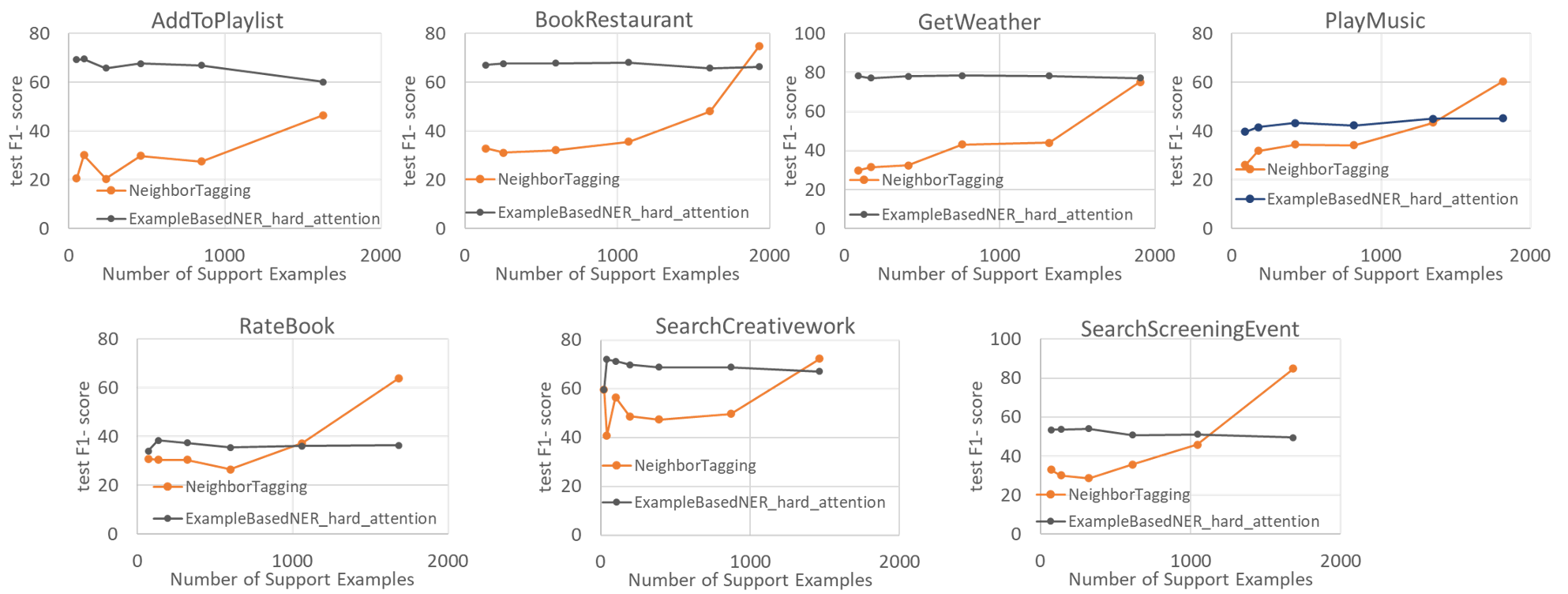}
\caption{Training on multiple domains and evaluating on a held-out domain: impact of number of support examples, comparison of our approach with Neighbor Tagging \citep{wiseman-stratos-2019-label}}
\label{fig:snips}
\end{figure*}

Another interesting question to answer is how much knowledge can we transfer from one domain to another domain from a similar distribution. 
In this set of experiments, we simulate a scenario where we combine different training sets coming from similar distributions 
and then evaluate the model on an similar but unseen target domain. 
We use the SNIPS datasets which has seven domains (AddToPlaylist, BookRestaurant, GetWeather, PlayMusic, RateBook, SearchCreativework, and SearchScreeningEvent). We train a model on a combined dataset from six domains and evaluate it on the remaining domain. Figure \ref{fig:snips} shows the results of such experiments in terms of the total number of support examples. Similar to tables \ref{tab:onto_2_other_main} and \ref{tab:conll_2_other_main}, we use 10, 20, 50, 100, 200, and 500 examples per entity type, but instead of showing the number of examples per entity type, we show the total number of examples in the figure.  For example, the first figure in Fig. \ref{fig:snips} shows the experiment where we combine the six domains of BookRestaurant, GetWeather, PlayMusic, RateBook, SearchCreativework, and SearchScreeningEvent to train a model and then evaluate it on the held-out domain of AddToPlaylist. Based on this figure, we observe that our approach is consistently and significantly better than the baseline when the number of example per entity is up to 100. For the extreme scenario when the number of the example per entity is 200 or 500, these two methods are comparable.  

\subsubsection{Performance on Our Proprietary Dataset}
\begin{figure}[h]
\centering
\includegraphics[width=0.4\textwidth,height=0.24\textwidth]{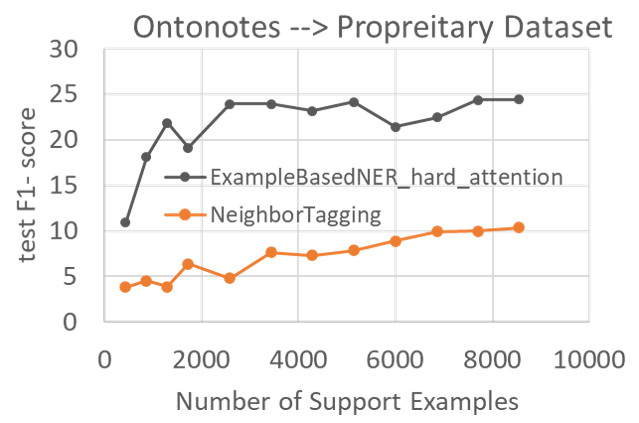}
\caption{Proprietary dataset scoring analysis (trained a model on OntoNotes 5.0 and evaluated on the proprietary dataset)}
\label{fig:onto_2_proprietary_data}
\end{figure}

Finally, we also evaluate the proposed approach on our proprietary dataset. Figure \ref{fig:onto_2_proprietary_data} shows the performance of our approach compared with neighbor tagging on a model trained on the OntoNotes 5.0 dataset and evaluated on our proprietary dataset. Different samples of support examples are created by randomly sampling the entire support with a different number of examples per entity type (i.e., 5, 10, 15, 20, 30, 40, 50, 60, 70, 80, 90, 100). Similarly to previous results, our approach achieves a consistently better performance .

\section{Conclusion}
This paper presents a novel technique for train-free few-shot learning for the NER task. 
This entity-agnostic approach is able to leverage large open-domain NER datasets to learn a generic model and then immediately recognize unseen entities via a few supporting examples, without the need to further fine-tune the model. This brings dramatic advantage to the NER problem requiring quick adaptation and provides a feasible way for business users to easily customize their own business entities. To the best of our knowledge, this is the first work that applies the train-free example-based approach on the NER problem.  
Compared to the recent SOTA designed for this setting, i.e., the neighbor tagging approach, extensive experiments demonstrate that our proposed approach performs significantly better in multiple studies and experiments, especially for low number of support examples. 


\bibliographystyle{acl_natbib}

\clearpage
\appendix
\section{Appendix}

\subsection{Related Work}
Historically, the NER task was approached as a sequence labeling task with recurrent and convolutional neural networks, and then with the rise of the Transformer \citep{transformer}, pretrained Transformer-based models such as BERT \citep{devlin2018bert}. While the boundary is being pushed inch by inch on well-established NER datasets such as CONLL 2003 \citep{conll2003}, such large and well-labeled datasets can present an ideal and unrealistic setting. Often, these benchmarks have strong name regularity, high mention coverage, and contain sufficient training examples for the entity types thus providing sufficient context diversity \citep{lin2020rigorous}. This is called \textit{regular} NER. In contrast, \textit{open} NER does not have such advantages--entity types may not be grammatical and the training set may not fully cover all test set mentions. 

Robustness studies have been done with swapping out English entities with entities from other countries, such as Ethiopia, Nigeria, and the Philippines, finding drops of up to 10 points F1 leading to questions of whether current state-of-the-art models trained on standard English corpora are over-optimized, similar to models in the photography domain which have become accustomed to the perfect exposure of white skin (\citealp{agarwal2020entityswitched}). Other studies by \citealp{agarwal2020interpretability} expose that while context representations resulting from trained LSTM-CRF and BERT architectures contribute to system performance, the main factor driving high performance is learning name tokens explicitly, which is a weakness when it comes to open NER, when novel and unseen name tokens run abundant.

To deal with partial entity coverage, one approach is data augmentation. \citealp{zhi2020partially} compares strategies to combine partially-typed NER datasets with fully-typed ones, demonstrating that models trained with partially-typed annotations can have similar performance with those trained with the same amount of fully-typed annotations. In this theme of lacking datasets, one approach has been to utilize expert knowledge to craft labeling functions in order to create large distantly or weakly supervised datasets. \citealp{lison2020named} utilizes labeling functions to automatically annotate documents with named-entity labels followed by a trained hidden Markov model to unify the labels into a single probabilistic annotation. Another approach is to leverage noisy crowdsourced labels as well as pre-trained models from other domains as part of transfer learning--\citealp{simpson2020low} combines pre-trained sequence labelers using Bayesian sequence combination to improve sequence labeling in new domains with few labels or noisy data. To tackle noisy data resulting from distantly supervised methods, \citealp{ali2020fine} instead use an edge-weighted attentive graph convolution network that attends over corpus-level contextual clues, thus refines noisy mention representations learned over the distantly supervised data in contrast to just simply de-noising the data at the model's input. A new domain can also be a new language--\citealp{lee2020chinese} created a small labeled dataset via crowdsourcing and a large corpus via distant supervision for Chinese NER, leveraging mappings between English and Chinese and finding that pretraining on English data helped improve results on Chinese datasets.

Rather than data augmentation, another approach is few or zero-shot learning--relying on models pretrained on large datasets combined with a limited amount of support examples in the target domain. More generally for text classification, \citealp{meta-deng2020} disentangle task-agnostic and task-specific feature learning, leveraging large raw corpuses via unsupervised learning to first pretrain task-agnostic contextual features followed by meta-learning text classification. \citealp{Snell2017} also developed the prototypical network for classification scenarios with scarce labeled examples such that objects of one class are mapped to similar vectors.
For the NER task specifically, \citealp{few-shot2018} combined this model architecture with an RNN + CRF to simulate few-shot experiments on the OntoNotes 5.0 dataset.

\citealp{zhang2020mzet} use memory to transfer knowledge of seen coarse-grained entity types to unseen coarse and fine-grained entity types, additionally incorporating character, word, and context-level information combined with BERT to learn entity representations, as opposed to simply relying on similarity. Their approach relies on hierarchical structure between the coarse and fine-grained entity types.


Thinking more about the pretraining portion, given massive text corpora, language models like BERT \citealp{devlin2018bert} seem to be able to implicitly store world knowledge in the parameters of the neural network. \citealp{guu2020realm} use the masked language model pretraining task from BERT to pretrain prior to fine-tuning on the Open-QA task. \citealp{wiseman-stratos-2019-label} also use BERT as a pretrained language model for train-free few-shot learning, using the BIO (beginning-inside-out) format to classify each of the tokens independently while finetuning.

Another trend is formulating other NLP tasks as a machine reading comprehension (MRC) problem. For instance, \citep{wu2019coreference} and \citep{li2019unified} have done this for coreference resolution and NER, respectively. Using the MRC framework enables the combination of different datasets from different domains to better fine-tune. A challenge that re-framing NER as an MRC task helps to alleviate is when a token may be assigned several labels. For example, if a token is assigned two overlapping entities, this can be broken out into two independent questions \citep{li2019unified}.

In our approach, we utilize the MRC framework to fine-tune a BERT-based language model with a novel sentence-level attention and token-level similarity that is independent of training data entity types, achieving a better representation to perform few-shot evaluation in a new domain with superior results.

\subsection{Production Benefit of Example-Based Approach}
There are many benefits of the similarity and example-based approach. First, since the model is decoupled from the trained entity types, any change made to the entity types does not require retraining of the model. There are also many advantages of this in a production system. We can onboard a new customer without training any model.  For an existing running system, there are many non-developer editors working on the content to add or delete new entity types simultaneously; any of their changes can be immediately reflected in a production system without further training. 

Second, any prediction using this example-based approach can be traced back to  which example(s) contributed to the decision.  Every decision is interpretable based on examples and it is easy to produce online Key Performance Indicator (KPI) metrics for each example, thus enabling a natural way to measure the value of different content versions. Third, it is easy to debug and fix any dissatisfied (DSAT) cases. We can remove a bad example or add a new example to address any DSAT immediately. Meanwhile, this fix can also generalize to similar DSAT as well. This provides an easy way for content editors to improve their system independently and confidently without the involvement of either AI experts or model training. 

\subsection{Ablation Study}
In this section, we take a detailed look at our approach and investigate important factors of our model.

\subsubsection{Scoring Strategy}
One of the important factors in example-based NER is the scoring strategy, used to recognize the target entity type. Above, we explain our main approach as hard attention and provide some benchmark results. Potentially one could use soft attention in the scoring as well and we discuss two methods of using soft attention for scoring.

\textbf{Soft Attention Scoring:} This is very similar to hard attention but instead of using Eq. \ref{eq:15}, \ref{eq:16} in algorithm \href{alg:1}{I}, and \href{alg:2}{II} to calculate the probabilities, we use the following equations:

\begin{equation}
\label{eq:17}
P_i^{start} = \sum_{j=1}^{m_E} atten(q_{rep},   s_{rep}^{j})(q_i \odot s_{start}^{j})
\end{equation}
\begin{equation}
\label{eq:18}
P_i^{end} = \sum_{j=1}^{m_E} atten(q_{rep},   s_{rep}^{j})(q_i \odot s_{end}^{j})
\end{equation}

These two equations are similar to Eq. \ref{eq:7}, \ref{eq:8} in training but instead of using $K$, a fixed number of examples per entity type ($E$), we use all of the examples for that entity type, i.e., $m_E$.  

\textbf{Top $K$ Soft Attention Scoring:} Another approach is to use the sentence similarity scores (i.e., $atten$ scores) to filter the top $K$ sentences and measure the probabilities using those top $K$. In other words:

\begin{equation}
\label{eq:19}
P_i^{start} = \sum_{K \hspace{2pt} highest \hspace{2pt} atten} atten(q_{rep},   s_{rep}^{j})(q_i \odot s_{start}^{j})
\end{equation}
\begin{equation}
\label{eq:20}
P_i^{end} = \sum_{K \hspace{2pt} highest \hspace{2pt} atten} atten(q_{rep},   s_{rep}^{j})(q_i \odot s_{end}^{j})
\end{equation}

We also look into a heuristic approach as a very basic baseline. 

\textbf{Heuristic Voting Scoring:}  In the other scoring methods, we treat multiple examples all at once in an equation with different weights to calculate the score. A heuristic algorithm that we investigated is to treat each of the support examples separately and run span prediction per example per entity and then use voting between multiple examples of an entity type to produce the final predictions per entity type. Each support example gets an equal vote. Also, for the span score we use the base token-level similarity to measure the probability as shown in Eq. \ref{eq:5}, \ref{eq:6}. 
Algorithm \href{alg:3}{III} shows the voting algorithm. 

\begin{table}[h!]
\centering
\begin{tabular}{p{0.45\textwidth}}
\hline \textbf{Algorithm III: Entity Type Recognition - Voting Algorithm} \\
\hline
Suppose we have $M$ entity types with $m_l$ support examples per entity type and a query $Q$.\\
\hline
\textbf{for} each entity type $E$ in $E_1, .., E_M$ \textbf{do}:\\
$|$\hspace{10pt} \textbf{for} each support example $e$ in $m_E$ \textbf{do}:  \\
$|$\hspace{10pt} $|$ \hspace{10pt} get the predicted\\
$|$\hspace{10pt} $|$ \hspace{20pt}spans [$span_{i1}, span_{i2}, ..., span_{in}$] \\
$|$\hspace{10pt} $|$ \hspace{20pt}using algorithm \href{alg:4}{IV}\\
$|$\hspace{10pt} \textbf{get} the final prediction per entity as top \\
$|$\hspace{3pt}voted spans of  [$span_{i1}, span_{21}, ..., span_{{m_E}n}$]  \\
\\
\textbf{aggregate} all the predictions per entity type as final output\\
\hline
\end{tabular}
\label{alg:3}
\end{table}

In this algorithm, we treat each entity type separately and use the support  examples of that entity type to recognize the spans for that specific entity type. In the end, we accumulate all of the recognized spans for different entity types and take all of them as final prediction. For example, let us assume that we have $M$ entity types where each has $m_l$ ($l$ in $1,2,...,M$) support examples. Note that one example can have different entity types, from which we create multiple support examples from that example where each support example has only one entity. Now, for a query $Q$, and entity type $E$ with $m_E$ support examples, we get the top $n$ predicted spans ($PS$) for each support example ($e$) of the $E$ using algorithm \href{alg:4}{IV}. 

\begin{table}[h!]\label{alg:4}
\centering
\begin{tabular}{p{0.45\textwidth}}
\hline \textbf{Algorithm IV: Choose top $n$ best spans} \\
\hline
$start_{indexes}$=get $n$ top indexes from top P(start)   \\
$end_{indexes}$ = get $n$ top indexes from top P(end)   \\
\textbf{for} $start_{id}$ in $start_{indexes}$ \textbf{do}: \\
$|$\hspace{10pt} \textbf{for} $end_{id}$ in $end_{indexes}$ \textbf{do}:  \\
$|$ \hspace{10pt} $|$ \hspace{10pt} \textbf{if} $start_{id}$ , $end_{id}$ in the query \\
$|$ \hspace{10pt} $|$ \hspace{15pt} $\&$ $start_{id}$ $<$ $end_{id}$:  \\
$|$ \hspace{10pt} $|$ \hspace{10pt} $|$ \hspace{10pt} calculate $start_{prob}, end_{prob}$ \\
$|$ \hspace{10pt} $|$ \hspace{40pt}using Eq. \ref{eq:5}, \ref{eq:6}\\
$|$ \hspace{10pt} $|$ \hspace{10pt} $|$ \hspace{10pt} add ($start_{id}$, $end_{id}$, \\
$|$ \hspace{11pt} $|$ \hspace{40pt} $start_{prob}$, $end_{prob}$) to the output \\
\\
sort the spans based on the ($start_{prob}$ + $end_{prob}$) to select the top $n$ high probable spans \\
\hline
\end{tabular}
\end{table}

In short, we select the top start tokens and end tokens based on the probabilities, build the valid spans, sort them based on the highest summation of start probability and end probability, and select the top $n$ spans (i.e., [$span_1$, $span_2$, ..., $span_n$]).  :

\[ PredSpans_{e_1^E}^Q = [span_{11}, span_{12}, ..., span_{1n}] \]
\[ PredSpans_{e_2^E}^Q = [span_{21}, span_{22}, ..., span_{2n}] \]
\[ ... \]
\[ PredSpans_{e_{m_E}^E}^Q = [span_{{m_E}1}, ..., span_{{m_E}n}] \]

To finalize the span prediction for an entity type $E$, we take the vote on the top predicted span from each support example of $E$, i.e.:

\[ Pred_E^Q=[TopVoted (span_{11}, span_{21},...,span_{{m_E}1})] \]

And finally, we aggregate the predictions of different entity types in the query $Q$ to get the final result as:

\[ Pred^Q = \cup_{i=1}^M (Pred_{E_i}^Q) \]

It should be noted that in this method, the predicted spans of different entity types could potentially have overlaps. The voting algorithm is a simple heuristic approach with some gains and drawbacks that we will discuss. 

\begin{figure*}[h!]
\centering
\includegraphics[width=\textwidth,height=0.2\textwidth]{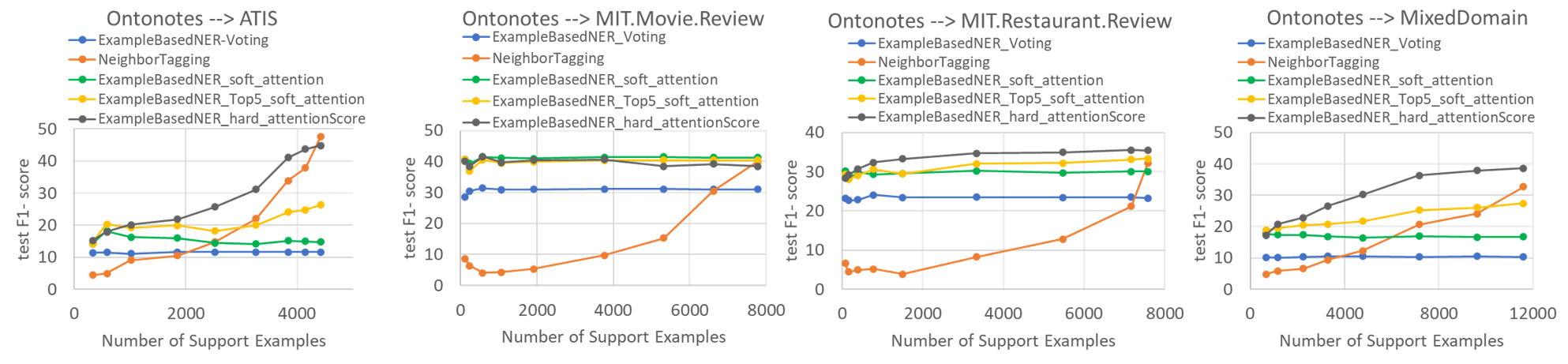}
\caption{Impact of number of support examples, comparison of different scoring algorithms}
\label{fig:onto_ablation_study}
\end{figure*}

To compare the performance of these new scoring algorithms, we run a benchmark study as before. Figure \ref{fig:onto_ablation_study} shows the results of transferring knowledge from OntoNotes 5.0 to other specific domains.

In general, the hard attention scoring algorithm achieves the best performance on a different number of examples. Note that the voting algorithm performs almost the same regardless of the number of support examples. This is because in this algorithm, we don't consider the scores of the entities from different examples and treat them as equally important. For instance if we have five examples, the voting result could be {`tag1':3, `tag2':2}, resulting in a prediction of `tag1' as it has the most votes. Now, if we increase the number of examples, we could have {`tag1':5, `tag2':4, `tag3':1} but still the resulting prediction is `tag1'. As described in this algorithm the actual prediction score for each of the support examples has not been taken into account. 

Similarly, the behavior of soft attention scoring does not change much in relation to the number of support examples. This is due to the fact that the summation in eq. \ref{eq:17}, \ref{eq:18} plays the role of an averaging window, and as we have increased the size of the window to all support examples, the aggregated results remain unchanged. In contrast, as we have a smaller window size of $K=5$ in top-K soft attention approach, we observe variations when changing the number of examples. However, the performance of this top-K soft attention is not as good as neighbor tagging when confronted with a higher number of examples. This gap is mitigated when using hard attention. 

For hard attention, we also tried different $K$ values for top $K$, and although the results were comparable, all cases with $K=5$ which is used during training, gave the best performance.

\begin{figure}[h!]
\centering
\includegraphics[width=0.4\textwidth,height=0.6\textwidth]{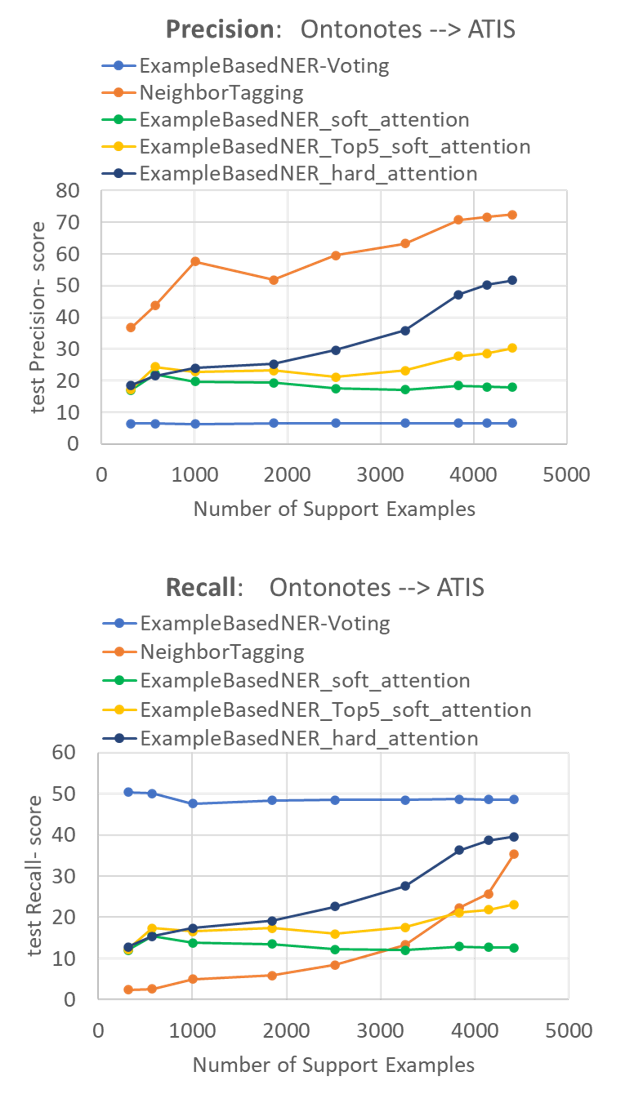}
\caption{Precision and recall analysis}
\label{fig:onto_precision_recall_ablation}
\end{figure}

To analyze the detail of each algorithm, we take a look at the precision and recall of different approaches. Figure \ref{fig:onto_precision_recall_ablation} shows the precision and recall of different approaches from a model trained on the OntoNotes 5.0 dataset and evaluated on the ATIS test set using a different number of ATIS support examples. Based on this figure, it is clear that the voting algorithm tends to have higher recall but low precision. In contrast, neighbor tagging tends to achieve higher precision than recall. Meanwhile the example-based NER approach with hard attention achieves a balance between precision and recall. 

\subsection{Negative Results}
We briefly describe a few ideas that did not look promising in our initial experiments:
\begin{itemize}
    \item We initially attempted to use entity masking when pre-training the BERT language model (i.e., masking the entities in the training set and pre-training the model to guess the entities) and then applying the fine-tuning approach. Although this approach helped a little bit on the seen entity type recognition, it was not helpful for train-free few-shot learning of unseen entity types. 
    \item Similar to neighbor tagging, we also applied the BIO tagging prediction as a scoring function on top of our fine-tuned model, but the results were not as good as our attention-based approach especially with a low number of support examples.
    \item Similar to neighbor tagging, we tried to run the prediction at word-level rather than token-level (since the BERT wordpiece tokenizer splits the words into multiple tokens and neighbor tagging uses the first token as the word representative and applies word-level prediction). The results were better than our heuristic voting algorithm but not as good as the current token-level scoring structure.
    \item We also experimented with the prototypical network structure using BERT language model but did not get comparable results.
    \item We also tried different attention mechanisms as well as similarity measures in our training and scoring approach but found that they had lower performance.
    
\end{itemize}

\end{document}